\documentclass[10pt,twocolumn,letterpaper]{article}

\usepackage[pagenumbers]{wacv}

\usepackage{graphicx}
\usepackage{amsmath}
\usepackage{amssymb}
\usepackage{booktabs}
\usepackage{tabularx}
\usepackage{adjustbox}
\usepackage{multirow}
\usepackage{float}
\usepackage{placeins}
\usepackage{colortbl}
\usepackage{array}
\usepackage{framed}
\usepackage{stfloats}
\usepackage{pifont}
\usepackage{algorithm}
\usepackage{algpseudocode}
\usepackage{listings}

\usepackage[table,xcdraw,dvipsnames]{xcolor}

\definecolor{darkgreen}{HTML}{006400}
\definecolor{beforeAfterBG}{HTML}{CCFFCC}
\definecolor{lightGreen}{HTML}{D4FCDC}
\definecolor{mediumGreen}{HTML}{A1E99A}
\definecolor{darkGreen}{HTML}{66CC66}
\definecolor{lightRed}{HTML}{FFCCCC}

\newcommand{\greenTick}{\textcolor{darkgreen}{\checkmark}}
\newcommand{\redCross}{\textcolor{red}{$\times$}}

\usepackage{tcolorbox}
\tcbuselibrary{skins,breakable,listings}

\lstdefinelanguage{prompt}{
  keywords      = {SYSTEM,USER,TASK,TAXONOMY,EXAMPLES,BEGIN},
  sensitive     = true,
  keywordstyle  = \color{RoyalBlue3}\bfseries,
  morecomment   = [l]{##},
  commentstyle  = \itshape\color{DarkOliveGreen4},
}
\lstdefinestyle{promptstyle}{
  language         = prompt,
  basicstyle       = \ttfamily\footnotesize\linespread{0.92}\selectfont,
  columns          = fullflexible,
  breaklines       = true,
  keepspaces       = true,
  showstringspaces = false,
}

\usepackage[pagebackref,breaklinks,colorlinks]{hyperref}

\usepackage[capitalize]{cleveref}
\crefname{section}{Sec.}{Secs.}
\Crefname{section}{Section}{Sections}
\Crefname{table}{Table}{Tables}
\crefname{table}{Tab.}{Tabs.}

\def\wacvPaperID{248} % *** Enter the WACV Paper ID here
\def\confName{WACV}
\def\confYear{2025}

\title{InterAct-Video: Reasoning‐Rich Video QA for Urban Traffic}

\author{
  Joseph Raj Vishal, Divesh Basina, Rutuja Patil, Manas Srinivas Gowda, Katha Naik, \\ Yezhou Yang, Bharatesh Chakravarthi \\
  Arizona State University}

\begin{document}
\twocolumn[{%
    \maketitle                % 1) print the title block

    % 2) your teaser figure — **not** a float any more
    \begin{center}
      \captionsetup{type=figure}  % tell LaTeX the next caption is a figure caption
      \includegraphics[width=.96\linewidth]{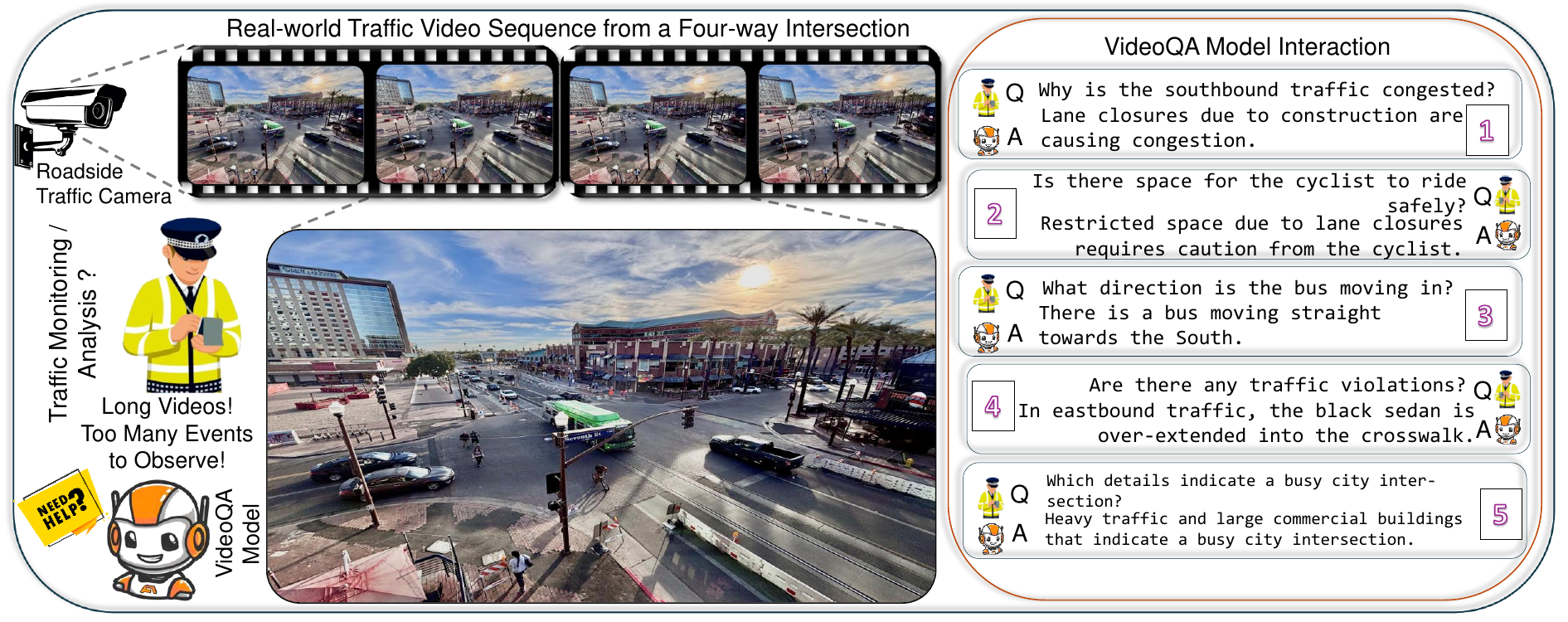}
      \captionof{figure}{\textbf{InterAct Video} analyzes a busy intersection and
        answers five kinds of questions: (1) Attribution, (2) Counting,
        (3) Event reasoning, (4) Reverse reasoning, and (5) Counter-factual
        inference.}
      \label{fig:teaser}
    \end{center}
    \vspace{-2ex}             % tighten the white space if you like
}]  

\begin{abstract}
Traffic monitoring is crucial for urban mobility, road safety, and intelligent transportation systems (ITS). Deep learning has advanced video-based traffic monitoring through video question answering (VideoQA) models, enabling structured insight extraction from traffic videos. However, existing VideoQA models struggle with the complexity of real-world traffic scenes, where multiple concurrent events unfold across spatiotemporal dimensions. To address these challenges, this paper introduces \textbf{InterAct VideoQA}, a curated dataset designed to benchmark and enhance VideoQA models for traffic monitoring tasks. The InterAct VideoQA dataset comprises 8 hours of real-world traffic footage collected from diverse intersections, segmented into 10-second video clips, with over 25,000 question-answer (QA) pairs covering spatiotemporal dynamics, vehicle interactions, incident detection, and other critical traffic attributes. State-of-the-art VideoQA models are evaluated on InterAct VideoQA, exposing challenges in reasoning over fine-grained spatiotemporal dependencies within complex traffic scenarios. Additionally, fine-tuning these models on InterAct VideoQA yields notable performance improvements, demonstrating the necessity of domain-specific datasets for VideoQA. InterAct VideoQA is publicly available as a benchmark dataset to facilitate future research in 
real-world deployable VideoQA models for intelligent transportation systems. GitHub Repo: \url{https://github.com/joe-rabbit/InterAct_VideoQA}
\end{abstract}

\section{Introduction}
\label{sec:intro}

With rapid urbanization, efficient traffic monitoring has become a critical challenge for city planners and transportation authorities. Traditional approaches, such as sensor-based traffic monitoring \cite{10.1007/978-981-15-2188-1_17, Li2024}, manual video surveillance \cite{Haering2008}, and rule-based analytics \cite{YE2023108924,7838263}, often struggle to scale and provide deep insights into the complexities of real-world traffic dynamics. The rise of computer vision and deep learning has led to significant advancements in automated traffic analysis, enabling models to recognize vehicle trajectories \cite{ELLIOTT2019109, BHARILYA2024100733}, track pedestrian movement \cite{OHNO2024101914, herath2022adoption}, and detect incidents \cite{hamdi2020techniques, elsahly2022systematic, olugbade2022review,herath2022adoption}. However, these models focus mainly on object detection and event classification, lacking the ability to provide structured, interpretable, and context-sensitive insights from traffic video data \cite{qasemi2023traffic, xu2021sutd}. VideoQA has emerged as a powerful paradigm for structured video understanding, allowing AI models to generate meaningful responses to natural language queries based on video content \cite{bansal2019visual, shao2018find, wang2024omnivid}. This capability goes beyond conventional computer vision tasks, enabling models to reason about interactions, infer causality, and analyze dynamic scenarios within video sequences. While VideoQA has been widely explored in general-purpose datasets such as MovieQA \cite{song2024moviechat+, tapaswi2016movieqa}, ActivityNet \cite{yu2019activitynet}, TGIF-QA \cite{jang2017tgif}, and others, its application in traffic monitoring remains largely underexplored. Given the growing demand for intelligent transportation systems, the ability of VideoQA models to analyze traffic activity, understand spatiotemporal interactions, and extract actionable insights holds significant potential for enhancing real-time traffic analytics.

The integration of VideoQA into traffic monitoring has the potential to transform real-time decision-making and automate surveillance-based systems. One key application is incident detection and response \cite{xu2021sutd, zhou2025tumtraffic}, where AI models can reason over video sequences to detect incidents \cite{ameer2024traffic}, traffic violations \cite{ameer2024traffic}, and hazardous road conditions \cite{pena2020real}, shown in question $4$ of Figure \ref{fig:teaser}, enabling faster interventions by traffic authorities. A crucial application area is vulnerable road user (VRU) safety \cite{zoghlami20235g}, where VideoQA models can identify pedestrians, cyclists, and other non-motorized road users in complex intersection scenarios, portrayed in question $2$ of Figure \ref{fig:teaser}, aiding proactive safety measures.

Despite the potential of VideoQA in traffic monitoring, current AI-driven traffic analysis remains limited to object detection, trajectory tracking, and event classification \cite{li2022coda}. Widely used datasets such as Coda \cite{li2022coda}, UTD19 \cite{loder2020utd19}, and LargeST \cite{liu2023largestbenchmarkdatasetlargescale} provide benchmarks for tasks like object recognition and vehicle tracking but lack structured question-answer annotations that support VideoQA-based reasoning. In contrast, the previously mentioned general-purpose VideoQA datasets do not capture the complexities of traffic-specific scenarios, such as multi-vehicle interactions, occlusions, and reasoning over real-time events, as suggested by question 5 in Figure \ref{fig:teaser}. A fundamental challenge in applying VideoQA to traffic monitoring is the spatiotemporal complexity of real-world traffic scenes. State-of-the-art (SOTA) VideoQA models \cite{cheng2024videollama, li2024llava, wang2024qwen2vlenhancingvisionlanguagemodels} often struggle with this complexity, as existing datasets do not provide domain-specific benchmarks to evaluate their performance in traffic-related tasks, a major bottleneck being the absence of real-world data.

To bridge this gap, there is a pressing need for a well-annotated dataset, which would enable the systematic evaluation of AI models on traffic-specific reasoning tasks, support the development of interpretable AI-driven traffic monitoring systems, and facilitate research in real-time decision-making. Specifically, an ideal dataset should include spatiotemporal reasoning tasks, vehicle interaction analysis, incident detection, and context-aware question-answer pairs that reflect real-world traffic challenges. To address these limitations, this paper  \textbf{InterAct VideoQA} (\textbf{Inter}section \textbf{Act}ivity Video Question Answering), a novel benchmark dataset designed for VideoQA in traffic intersection monitoring. Specifically curated to facilitate the training and evaluation of AI models that can reason over real-world traffic scenarios and provide structured answers to complex queries. InterAct VideoQA comprises $8$ hours of real-world traffic footage collected from diverse intersection scenarios, segmented into $10$-second video clips to facilitate event-level understanding. The dataset includes over $25,000$ QA pairs, categorized into key reasoning tasks essential for traffic analysis.
%###
The study uses $8$ hours of traffic video, a modest volume given available sources, but even with this limited dataset, SOTA models struggle to generalize. The footages were carefully sampled to represent the most common traffic environments (urban arterials, residential streets, signalized intersections) and all key times of day (morning rush, midday lull, evening peak). Ongoing work is integrating additional sites and seasonal variations to broaden the dataset’s spatial and temporal coverage.
% ##
These tasks include spatiotemporal reasoning  \textit{“How long did it take the train to pass the intersection?”}, vehicle interaction analysis  \textit{“Did black sedan cut any vehicles off at the intersection?”}, incident detection  \textit{“Did any vehicles jump the signal?”}, traffic density estimation  \textit{“How is the traffic at the intersection?”}, road user behavior analysis \textit{“Does the black sedan show signs of impatience?”} and environmental context understanding \textit{ “Why does the person crossing the road in the video seem to be running?”}. By providing structured annotations tailored for traffic VideoQA, the dataset enables AI models to develop a deeper understanding of complex traffic scenarios, paving the way for advancements in intelligent transportation systems. Moreover, these annotations enhance model training, boost prediction accuracy, reduce incident risks, and optimize traffic flow management in urban environments. The key contributions of this work are summarized below.  \begin{itemize}
    \item \textbf{InterAct VideoQA Dataset}: A dedicated benchmark dataset for VideoQA in traffic monitoring, enabling structured reasoning over real-world traffic scenarios. 
    \item \textbf{Comprehensive Data Collection and Annotation}: The dataset includes $8$ hours of real-world traffic footage, segmented into $10$-second clips, with over $25,000$ QA pairs covering critical traffic attributes.
    \item \textbf{Evaluation of SOTA VideoQA Models}: The performance of existing VideoQA models is assessed on the dataset, highlighting challenges in spatiotemporal reasoning and complex traffic scene understanding.
    \item \textbf{Fine-Tuning and Performance Improvements}: This study also demonstrates the benefits of fine-tuning VideoQA models on traffic-related tasks, showcasing significant improvements in model accuracy and interoperability.
\end{itemize}

\begin{table*}[t]
\centering
\caption{Comparison of existing VideoQA datasets. \greenTick\ indicates presence, \redCross\ absence of a feature.}
\label{tab:table01}
\begin{adjustbox}{max width=\textwidth}
\begin{tabular}{c|c|c|c|c|c|c|c|c|c|c|c|c|c}
\toprule \toprule
\textbf{Type} &
  \textbf{Dataset} &
  \textbf{Year} &
  \textbf{Domain} &
  \textbf{Task Type} &
  \textbf{Annotation} &
  \textbf{\begin{tabular}[c]{@{}c@{}}Annotation \\ Density\end{tabular}} &
  \textbf{\begin{tabular}[c]{@{}c@{}}VideoQA \\ Pairs\end{tabular}} &
  \textbf{Reasoning} &
  \textbf{\begin{tabular}[c]{@{}c@{}}Color\\ Attributes\end{tabular}} &
  \textbf{\begin{tabular}[c]{@{}c@{}}Road\\ Signs\end{tabular}} &
  \textbf{\begin{tabular}[c]{@{}c@{}}Low\\ Resolution\end{tabular}} &
  \textbf{Intersection} &
  \textbf{Multi-Events} \\ \toprule \toprule
\multirow{5}{*}{\begin{tabular}[c]{@{}c@{}} {\rotatebox[origin=c]{90}{\textbf{Traditional}}} \end{tabular}} &
  STAR &
  2021 &
  Web &
  VideoQA &
  Manual &
  Moderate &
  22k/60K &
  \redCross &
  \redCross &
  \redCross &
  \redCross &
  \redCross &
  \redCross \\ 
 &
  EgoTaskVQA &
  2022 &
  Egocentric &
  VideoQA &
  Manual &
  Moderate &
  2k / 40k &
  \greenTick &
  \redCross &
  \redCross &
  \redCross &
  \redCross &
  \redCross \\ 
 &
  AGQA &
  2021 &
  YouTube &
  VideoQA &
  Template &
  Dense &
  9.6k / 192M &
  \greenTick &
  \redCross &
  \redCross &
  \redCross &
  \redCross &
  \greenTick \\ 
 &
  TVQA &
  2018 &
  TV Shows &
  VideoQA &
  Manual &
  Moderate &
  21.8k / 152.5k &
  \greenTick &
  \greenTick &
  \redCross &
  \redCross &
  \redCross &
  \redCross \\ 
 &
  MSRVTT-QA &
  2017 &
  Web Videos &
  VideoQA &
  Template &
  Sparse &
  10k / 243.7k &
  \redCross &
  \greenTick &
   \redCross &
   \redCross &
   \redCross &
   \redCross \\ \midrule
\multirow{10}{*}{\begin{tabular}[c]{@{}c@{}} {\rotatebox[origin=c]{90}{\textbf{Traffic}}}  \end{tabular}} &
  TUMTraffic-VideoQA &
  2025 &
  Roadside &
  VideoQA &
  Template &
  Sparse &
  $\sim1k/85k$ &
  \greenTick &
   \redCross &
   \redCross &
  \greenTick &
   \greenTick &
   \redCross \\ 
 &
  LingoQA &
  2024 &
  Driving &
  VideoQA &
  Manual &
  Moderate &
  28k/419k &
  \greenTick &
  \redCross &
  \redCross &
  \redCross &
  \redCross &
  \redCross \\ 
 &
  NuScenes-QA &
  2024 &
  Driving &
  ImageQA &
  Template + Manual &
  Sparse &
  850/460k &
  \redCross &
  \redCross &
  \redCross &
  \redCross &
  \redCross &
  \redCross \\ 
 &
  Drive-LM &
  2024 &
  Driving &
  ImageQA &
  Template + Manual &
  Moderate &
  188k/4.2M &
  \redCross &
  \redCross &
  \redCross &
  \redCross &
  \redCross &
  \redCross \\ 
 &
  City-3DQA &
  2024 &
  City &
  SceneQA &
  Manual &
  Moderate &
  193/450K &
  \greenTick &
  \redCross &
  \redCross &
  \greenTick &
  \redCross &
  \redCross \\ 
 &
  DRAMA &
  2023 &
  Driving &
  VideoQA &
  Manual &
  Moderate &
  18k/102k &
  \greenTick &
  \redCross &
  \redCross &
 \redCross &
  \redCross &
  \redCross \\ 
 &
  RoadTextVQA &
  2023 &
  Driving &
  VideoQA &
  Manual &
  Moderate &
  3.2k/10.5k &
  \redCross &
  \redCross &
  \greenTick &
  \redCross &
  \redCross &
  \redCross \\ 
 &
  Refer-KITTI &
  2023 &
  Driving &
  Referred-MOT &
  Manual &
  Moderate &
  18/- (818 grounding) &
  \redCross &
  \redCross &
  \redCross &
  \redCross &
  \redCross &
  \redCross \\ 
 &
  SUTD-TrafficQA &
  2021 &
  Wild Traffic &
  VideoQA &
  Manual &
  Moderate &
  10.8k/62.5k &
  \greenTick &
  \redCross &
  \redCross &
  \redCross &
  \greenTick &
  \redCross \\ 
\cmidrule{2-14} 
 &

 \textbf{InterAct VideoQA [Ours]} &
  2025 &
  Roadside &
  VideoQA &
  \begin{tabular}[c]{@{}c@{}}Dense Template \\ + Manual\end{tabular} &
  Dense &
  $\sim$2.9k/34.6k &
  \greenTick &
  \greenTick &
  \greenTick &
  \greenTick &
  \greenTick &
  \greenTick \\ \bottomrule \bottomrule
\end{tabular}
\end{adjustbox}
\end{table*}

%------------------------------------------------------------------------
\section{Related Study}
\label{sec:relatedStudy}

The section provides an overview of existing VideoQA as well as traffic VideoQA datasets that are currently available. It discusses the merits and drawbacks of these datasets in the context of VideoQA models designed for traffic-specific question answering. 

%-------------------------------------------------------------------------
\subsection{VideoQA Datasets}

Traditional datasets such as STAR \cite{wu2024star}, AGQA \cite{grunde2021agqa}, and EgoTaskVQA \cite{xue2023egocentric} have played a crucial role in advancing video action recognition. STAR focuses on identifying discrete real-world actions in short video segments, while AGQA provides a diverse set of questions targeting multi-event VideoQA. EgoTaskVQA extends these efforts by incorporating first-person perspectives, enhancing the understanding of egocentric activities. MSRVTT-QA \cite{xu2016msr} builds on the MSRVTT video collection by offering richly annotated question-answer pairs, while TVQA leverages TV show clips and subtitles to challenge models with contextual and temporal reasoning.

%-------------------------------------------------------------------------
\subsection{Traffic-related VideoQA Datasets}

In the domain of traffic-specific VideoQA datasets, several key contributions have advanced the field, including SUTDTrafficQA \cite{xu2021sutd}, DRAMA \cite{malla2023drama}, RoadTextVQA \cite{tom2023readinglanestextvideoqa}, LingoQA \cite{marcu2024lingoqa}, NuScenes-QA \cite{10.1609/aaai.v38i5.28253}, and TUMTraffic \cite{zhou2025tumtraffic}. SUTDTrafficQA focuses on causal reasoning in unconstrained traffic scenarios but emphasizes isolated events rather than the continuous, overlapping interactions typical of real-world traffic. DRAMA addresses open-ended driving instructions, capturing dynamic and context-driven behavior, while RoadTextVQA specializes in road text and signage recognition, which are essential for traffic understanding but are limited to capturing dynamic events.

LingoQA and NuScenes-QA improve traffic-related VideoQA, while TUMTraffic provides a roadside perspective but lacks intersection-specific data, making it less effective for analyzing critical traffic junctures. Furthermore, Chaotic World \cite{ong2023chaotic} explores crowd management and emergency response, focusing on extreme and unpredictable events in disaster scenarios. These datasets have significantly advanced the understanding of crowd behavior, event recognition, disaster prediction, and crisis management in traffic. Despite these advancements, there remains a gap in datasets that capture dense, asynchronously occurring multi-event interactions, reflecting the full complexity of real-world traffic as shown in the Table  \ref{tab:table01}. To address this, we introduce InterAct VideoQA, a benchmark dataset designed for video question answering in traffic intersection monitoring. By offering an intersection-focused perspective with dense, continuous, and overlapping interactions, InterAct VideoQA bridges this gap, enabling a more comprehensive analysis of traffic dynamics.

\begin{figure*}[t]
  \centering
\includegraphics[width=0.86\linewidth]{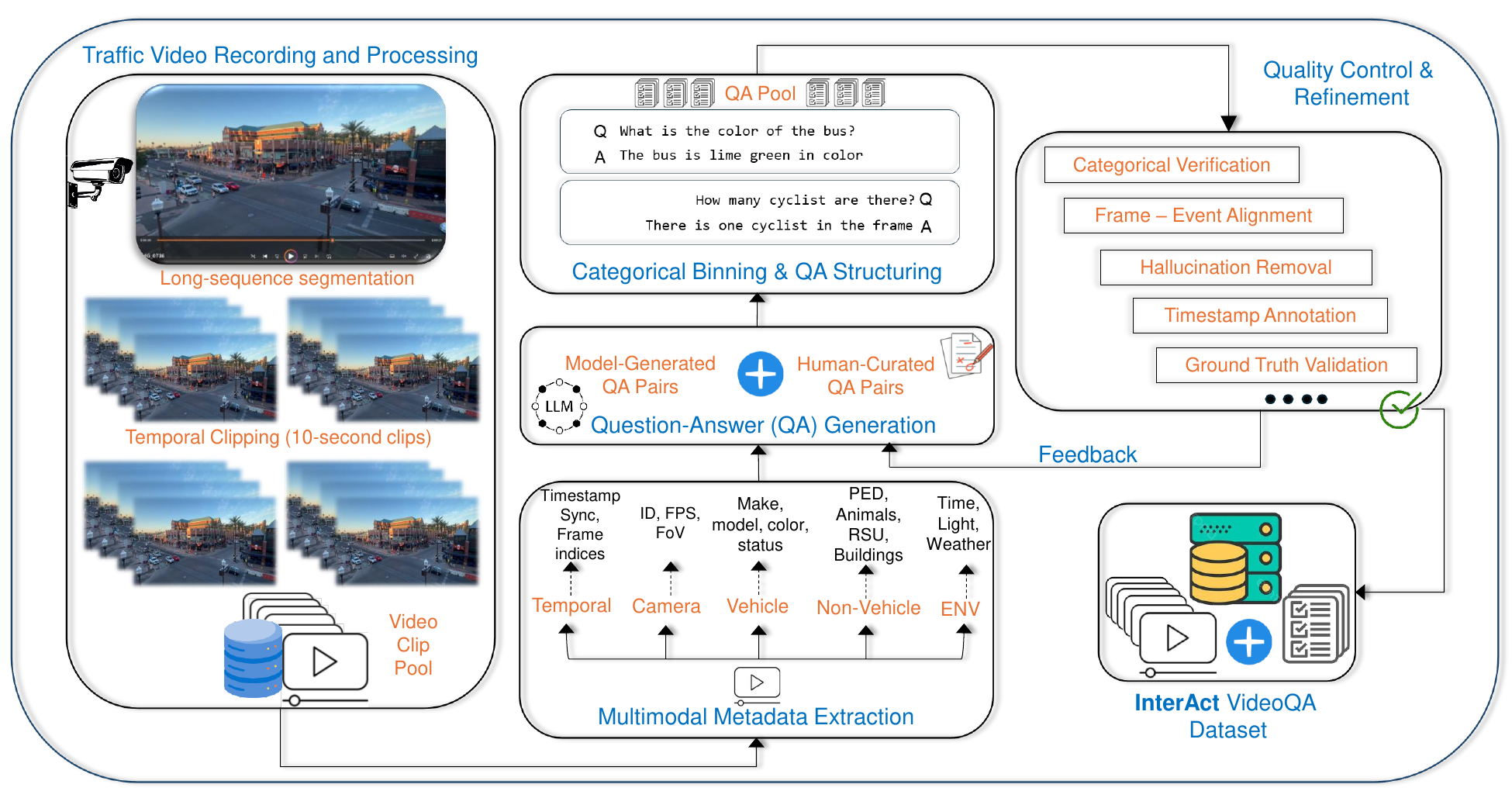}
\caption{An overview of the InterAct VideoQA pipeline for creating a high-quality question-answer dataset from traffic videos.
}
\label{fig02}
\end{figure*}

\section{InterAct VideoQA}
\label{sec:InterActVideoQA}

InterAct VideoQA is a traffic intersection monitoring dataset designed to capture the complex and chaotic dynamics of intersections in densely populated areas with varying pedestrian and traffic densities. 
InterAct VideoQA supports the development and evaluation of ITS, improving their overall performance in real-world scenarios.

%-------------------------------------------------------------------------
\subsection{Dataset Overview}

InterAct VideoQA records and examines traffic dynamics at intersections under diverse environmental conditions, varying pedestrian and vehicular densities, and typical urban settings. The dataset comprises $8$ hours of traffic footage segmented into $10$-second clips. The study limits each segment to $10$-seconds to balance annotation effort against dynamic detail within any shorter window, virtually no change in traffic flow occurs, while longer clips introduce an overwhelming number of interactions. Consequently, a $10$-second window captures sufficient dynamics for the study purposes without overloading the annotators. Offering detailed annotations and reasoning categories, including attribution, counting, reverse reasoning, event reasoning, and counterfactual inference, making it an invaluable resource for event-level insights, multi-event tracking, spatio-temporal analysis, event forecasting, and inference.

\subsection{Data Collection}
Data for InterAct-VideoQA was recorded at multiple intersections in urban setting, near the university campus. This location offered a diverse range of pedestrian and vehicular densities, from the vibrant energy of the district’s nightlife to the heavy congestion typical of interstate rush hours. Data collection employed mounted cameras and mobile devices, ensuring comprehensive coverage across different times of day (morning, afternoon, and night) and varied weather conditions. The methodology also captured urban scenarios such as road construction, closures, and emergency maneuvers, accurately reflecting real-world traffic dynamics.

%-------------------------------------------------------------------------
\subsection{Data Annotation}

The InterAct VideoQA annotation framework, illustrated in Figure \ref{fig02}, integrates traffic video recording and processing with a hybrid approach, combining manual labeling and automated assistance from GPT \cite{openai2024gpt4ocard}. This method leverages both human expertise and AI-driven efficiency to enhance annotation accuracy.
The eight-hour traffic footage is first segmented into 10-second clips, forming a structured pool of video data. Each clip is further broken down into individual frames, enabling high-resolution frame-level analysis using GPT to perform multimodal extraction of metadata, as shown in the traffic video recording and processing of Figure \ref{fig02}. This extracted metadata captures key traffic elements, including vehicle attributes, movement patterns, environmental conditions, and pedestrian metadata.

Following metadata extraction, QA generation and curation take place. The QA pairs are systematically binned into categories to ensure a structured representation of the dataset, using both human annotation and generative models, as illustrated in Figure \ref{fig02}. The question types include attribute recognition, which identifies key characteristics of objects in the scene such as vehicle type, color, or traffic signals, counting, which determines numerical values such as the number of cars, pedestrians, or traffic violations in a given frame, event reasoning, which involves understanding causal relationships between observed events, such as predicting the outcome of a red-light violation, reverse reasoning, which analyzes hypothetical scenarios by asking ``what if'' questions, such as how traffic flow would change if a signal were different and counterfactual inference, which constructs negated or inverted questions to mitigate model hallucinations and assess logical consistency. These question types are adapted from SUTD-TrafficVideoQA \cite{xu2021sutd} to ensure robust analytical coverage.
Once the question-answer structuring is completed, human annotators verify and refine the labels to maintain accuracy and consistency. The annotated dataset is then subjected to quality control and refinement, as detailed in Section \ref{sec:DatasetValidation}, to uphold annotation reliability and ensure its effectiveness for traffic analysis research.

\begin{figure*}[th]
  \centering
\includegraphics[width=0.92\linewidth]{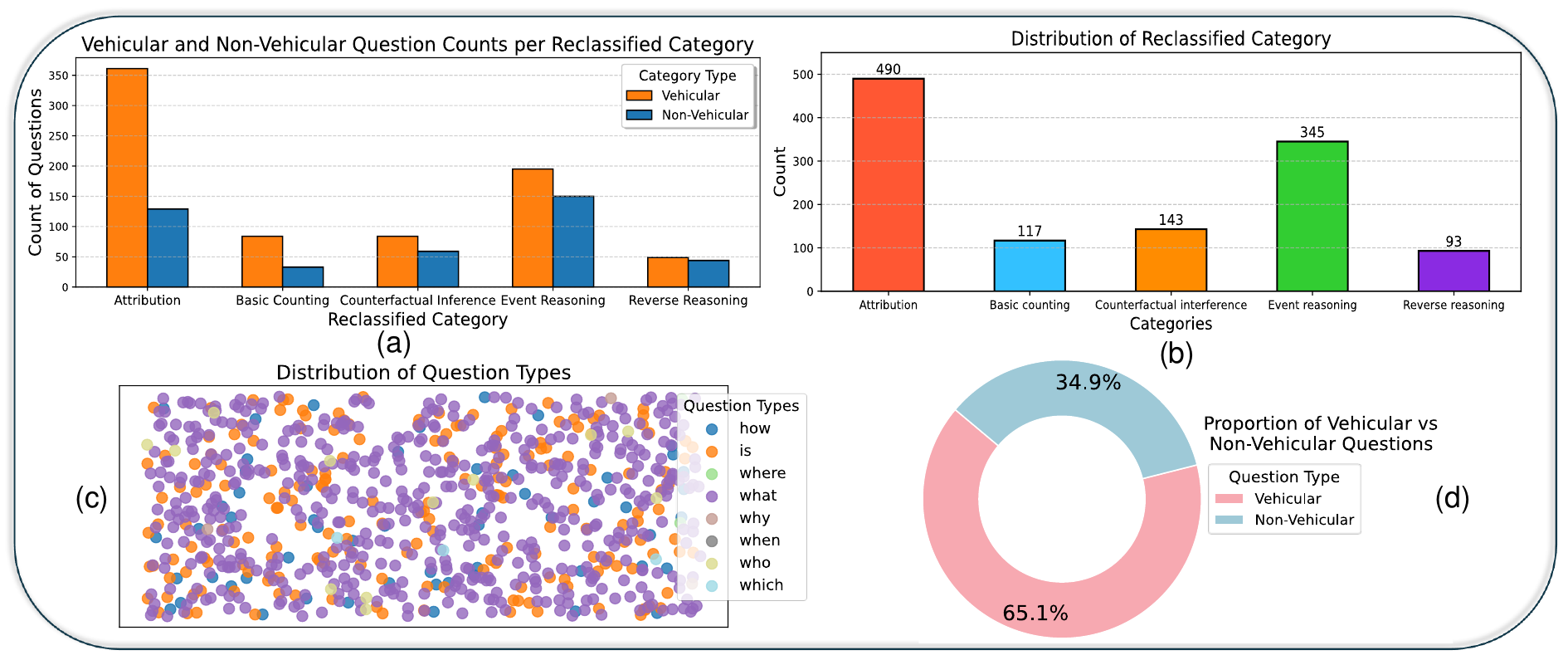}
\caption{Question distribution – (a) Vehicular vs. non-vehicular question counts by category, (b) Overall distribution of question types, (c) question type distribution, and (d) Overall proportion of vehicular vs. non-vehicular questions.}
\label{fig03}
\end{figure*}

\subsection{Dataset Structure and Statistics}
The dataset comprises $28,800$ question-answer pairs, distributed across various reasoning categories with a higher concentration in counting, attribute recognition, and event reasoning, followed by counterfactual inference and reverse reasoning, as shown in Figure \ref{fig03}(a). 
The emphasis of the dataset on vehicular related questions and the predominance of categories of attribution and event reasoning are illustrated in Figures \ref{fig03}(b), while Figure \ref{fig03}(c) shows the distribution of questions based on  \textit{“what”},  \textit{“where”}, and  \textit{“how”}. Figure \ref{fig03}(d) shows the overall proportion of vehicular and non-vehicular questions. Event reasoning, which involves understanding the relationships between different events, plays a crucial role in interpreting overlapping or concurrent events in real-world scenarios. The counterfactual inference category further indicates an effort to reason about alternative outcomes, which is often necessary when multiple events interact.
The dominance of \textit{“What”} and \textit{“Is”} questions, shown in Figure \ref{fig03}(c), suggests a preference for descriptive and confirmatory inquiries, which are essential for breaking down complex, multi-event situations into manageable components. The higher proportion of vehicular-related questions implies that these multi-event scenarios often involve traffic dynamics, such as interactions between cars, pedestrians, and cyclists. This structured questioning helps disentangle multiple events and understand their impact on each other, which is critical for traffic analysis, incident investigations, and decision-making about autonomous vehicles.
Event reasoning questions such as \textit{“At 0:09, was there a white sedan turning at the intersection?”} incorporate spatio-temporal queries, requiring models to infer interactions over time. Whereas reverse reasoning questions, such as \textit{“Did the cyclist overtake the red sedan at signal 2?”}, focus on simultaneous events and interaction-based inference. Meanwhile, counterfactual questions, such as \textit{“Was there an orange bus?”}, to test the capabilities of the model when events do not occur, and to hallucinate events.

\section{Dataset Validation}
\label{sec:DatasetValidation}

\subsection{Human Evaluation of InterAct VideoQA}
The dataset was annotated by human annotators assisted by the GPT-o3 model. Following this, it was manually verified by human evaluators. A rigorous evaluation was conducted for each video and its corresponding annotations to eliminate inconsistencies from both human and GPT-generated responses to ensure the highest level of consistency and reliability. 

During this evaluation, it was found that a significant number of GPT-generated responses contained hallucinated, factually incorrect, or fabricated information that did not align with the corresponding video content. To overcome bias in the dataset towards the questions, we introduced a human-evaluation rubric based on relevance, correctness, and completeness. We employed undergraduates to manually assess each question by splitting the dataset at random and assessing each video. Therefore, human evaluators cross-referenced every question-answer pair against its video footage. Problematic question-answer pairs were either discarded or manually corrected to ensure the objectiveness of the ground truth. Although labor-intensive, this was an essential part of the quality control pipeline to maintain the credibility and usability of the dataset for training robust AI models. Every clip released in InterAct videoQA is fully anonymized. 

% The workflow—approved by the \textbf{Arizona State University IRB}. See the appendix for the workflow.

\begin{table*}[ht]
\centering
\caption{Benchmark performance metrics (\%) of models across various question types.}
\label{tab:table02}
\resizebox{1\linewidth}{!}{%
\begin{tabular}{c|c|ccccc|ccccc|ccccc}
\toprule \toprule
\multirow{2}{*}{\textbf{FineTuning}} &
\multirow{2}{*}{\textbf{Questions}} & 
  \multicolumn{5}{c|}{\textbf{VideoLlama2}} &
  \multicolumn{5}{c|}{\textbf{Llava-NexT-Video}} &
  \multicolumn{5}{c}{\textbf{Qwen2-VL-7B-hf}} \\ \cline{3-17} 
 &
   &
  BLEU &
  ROUGE &
  METEOR &
  CIDEr &
  SPICE &
  BLUE &
  ROUGE &
  METEOR &
  CIDEr &
  SPICE &
  BLUE &
  ROUGE &
  METEOR &
  CIDEr &
  SPICE \\ \toprule \toprule

\multirow{5}{*}{\rotatebox[origin=c]{90}{\textbf{After}}} &
Basic Counting &
  - & 27.78 & 29.82 & \cellcolor{lightGreen}117.52 & 32.50 &
  - & \cellcolor{lightRed}10.53 & \cellcolor{lightRed}22.55 & \cellcolor{lightRed}62.04 & \cellcolor{lightRed}4.35 &
  - & \cellcolor{darkGreen}27.78 & \cellcolor{lightRed}29.82 & \cellcolor{darkGreen}117.52 & 32.50 \\ \cline{2-17} 
& Attribution &
  \cellcolor{lightGreen}1.08 & \cellcolor{lightGreen}26.01 & \cellcolor{lightRed}30.59 & \cellcolor{lightRed}84.41 & \cellcolor{lightRed}24.25 &
  - & \cellcolor{lightRed}26.27 & \cellcolor{lightRed}34.06 & \cellcolor{lightRed}91.24 & \cellcolor{lightRed}23.83 &
  \cellcolor{mediumGreen}2.70 & \cellcolor{darkGreen}27.48 & \cellcolor{lightRed}36.09 & \cellcolor{lightRed}108.44 & \cellcolor{mediumGreen}27.16 \\ \cline{2-17} 
 & Event Reasoning &
  15.15 & \cellcolor{darkGreen}37.70 & \cellcolor{mediumGreen}51.61 & \cellcolor{darkGreen}279.44 & \cellcolor{mediumGreen}34.30 &
  \cellcolor{mediumGreen}15.15 & \cellcolor{mediumGreen}36.14 & \cellcolor{mediumGreen}50.00 & \cellcolor{darkGreen}298.99 & \cellcolor{mediumGreen}34.40 &
  \cellcolor{darkGreen}10.14 & \cellcolor{darkGreen}34.94 & \cellcolor{darkGreen}45.45 & \cellcolor{darkGreen}250.95 & \cellcolor{darkGreen}31.04 \\ \cline{2-17} 
& CounterFactual &
  - & 31.25 & 31.20 & - & 18.18 &
  - & 31.25 & 31.20 & - & 18.18 &
  - & 31.25 & 38.75 & - & 19.05 \\ \cline{2-17} 
& Reverse Reasoning &
  2.65 & \cellcolor{lightGreen}24.21 & \cellcolor{lightGreen}38.22 & \cellcolor{darkGreen}136.53 & \cellcolor{mediumGreen}23.98 &
  7.39 & \cellcolor{mediumGreen}31.24 & \cellcolor{mediumGreen}39.57 & \cellcolor{mediumGreen}182.07 & \cellcolor{lightRed}22.25 &
  \cellcolor{darkGreen}16.83 & \cellcolor{darkGreen}28.44 & \cellcolor{mediumGreen}39.03 & \cellcolor{darkGreen}253.70 & \cellcolor{darkGreen}23.82 \\ \midrule \midrule

\multirow{5}{*}{\rotatebox[origin=c]{90}{\textbf{Before}}}
 &
Basic Counting &
  - & 27.78 & 29.82 & 117.20 & 32.50 &
  - & 27.78 & 29.82 & 117.20 & 32.50 &
  - & 16.89 & 37.42 & 109.07 & 32.50 \\ \cline{2-17} 
& Attribution &
  0.89 & 25.66 & 32.54 & 87.96 & 24.40 &
  2.70 & 27.73 & 35.59 & 104.20 & 24.20 &
  1.67 & 14.95 & 47.35 & 148.11 & 24.46 \\ \cline{2-17} 
& Event Reasoning &
  - & 12.57 & 47.66 & 241.75 & 29.98 &
  10.43 & 33.68 & 45.89 & 235.47 & 29.98 &
  0.64 & 12.52 & 22.05 & 74.29 & 8.08 \\ \cline{2-17} 
& CounterFactual &
  - & 31.25 & 31.20 & - & 18.18 &
  - & 31.25 & 31.20 & - & 18.18 &
  - & 28.00 & 31.32 & - & 20.00 \\ \cline{2-17} 
& Reverse Reasoning &

  2.65 & 23.74 & 37.33 & 130.88 & 22.25 &
  7.39 & 27.23 & 37.61 & 178.02 & 23.81 &
  0.77 & 14.95 & 37.94 & 130.25 & 5.88 \\  \bottomrule  \bottomrule
\end{tabular}%
}
\end{table*}

 %------------------------------------------------------------------------

 \section{Experiments}
\label{experiments}
 Three SOTA multi-modal models-VideoLLaMA2 \cite{cheng2024videollama2advancingspatialtemporal}, Qwen2-VL-7B-Instruct \cite{Qwen2VL, Qwen-VL}, and LLaVA-NeXT-Video \cite{zhang2024llavanextvideo,liu2024llavanext,liu2023improvedllava,liu2023llava} were evaluated to assess their performance and robustness in video understanding tasks. The models were selected based on a previous study conducted \cite{vishal2024eyes}. The section is divided into three parts: model descriptions, performance, and robustness analysis.

 \subsection{Model Descriptions}
\textbf{VideoLLaMA2} is designed to enhance spatial-temporal modeling and audio understanding in video-language tasks. It incorporates a spatio-temporal convolution connector to capture the dynamics of video data effectively. Additionally, an audio branch is integrated through joint training, enriching the model's multi-modal understanding capabilities by incorporating audio cues.

\textbf{Qwen2-VL-7B-Instruct}, built upon the Qwen2 language architecture, Qwen2-VL-7B-Instruct, a multi-modal model, processes up to $64$ video frames and is trained on the LLaVA-Video-178K and LLaVA-OneVision datasets. It is designed to interact with images, multi-image inputs, and videos, enhancing its ability to comprehend and generate language grounded in visual content.

\textbf{LLaVA-NeXT-Video}, an open-source chatbot fine-tuned on multi-modal instruction-following data, LLaVA-NeXT-Video is built upon the LLaVA-NeXT framework. It is trained on a mix of video and image data, processing videos by sampling $32$ frames per clip uniformly, to achieve better video understanding capabilities.

\begin{figure*}[t]
  \centering
\includegraphics[width=0.92\linewidth]{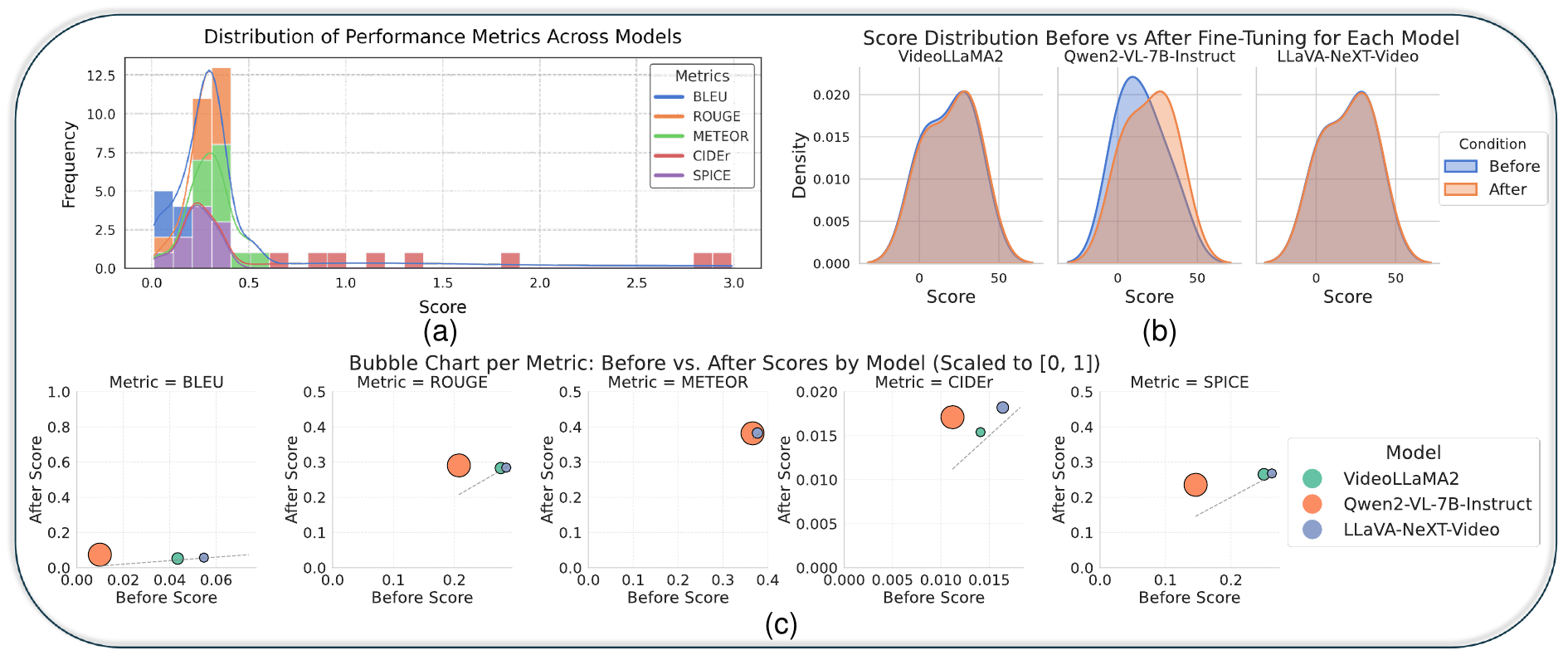}
\caption{Analysis of (a) performance score distributions, (b) pre/post-fine-tuning comparisons, and (c) multi-metric improvements.}
\label{fig04}
\end{figure*}

\subsection{Performance Analysis}

The study evaluated VideoLlama2, Llava-Next-Video, and Qwen2-VL-7B-hf using standard natural language processing metrics: BLEU \cite{papineni2002bleu}, ROUGE \cite{lin2004rouge}, METEOR \cite{banerjee2005meteor}, CIDEr \cite{vedantam2015cider}, and SPICE \cite{anderson2016spice}. These models were selected based on their architecture and promising results during a preliminary study \cite{vishal2024eyesroadstateoftheartvideo}. These metrics were chosen as they are standard evaluations, which allow direct comparison with annotated text. Individually, BLEU and ROGUE cover lexical overlap, METEOR covers synonym-aware matching, CIDEr focuses on detail weighting, and SPICE focuses on scene graph semantics. The study supplements the scores with human checks on causality and spatial accuracy, which confirm that the high metric scoring descriptions genuinely reflect correct scene understanding. While we agree that there exist and future benchmarks may incorporate deeper reasoning probes, at present these five metrics remain rigorous, reproducible, and accepted evaluations.

The detailed analysis in Table~\ref{tab:table02} and Figure~\ref{fig04} reveals distinct patterns between different types of questions after fine-tuning. In basic counting tasks, Qwen2-VL-7B-hf and VideoLlama2 achieved identical performance with CIDEr scores of $118$, while Llava-NexT-Video showed comparatively lower performance. For attribution questions, Qwen2-VL-7B-hf demonstrated superior performance with ROUGE  and CIDEr scores, marking a significant improvement from its pre-fine-tuning state.

As shown in Table~\ref{tab:performance_improvements}, the fine-tuning process yielded significant improvements across models, with Qwen2-VL-7B-Instruct showing the most substantial gains. Qwen2-VL-7B-Instruct demonstrated remarkable improvements in BLEU scores and SPICE metrics, indicating enhanced precision and semantic understanding. While LLaVA-NeXT-Video showed modest improvements across most metrics, it experienced a slight decline in ROUGE scores.

Event reasoning emerged as a strength across all models post-fine-tuning. In reverse reasoning, Qwen2-VL-7B-hf showed the most dramatic enhancement, achieving the highest BLUE score and CIDEr score.
The model demonstrated remarkable consistency in counterfactual question-based tasks. VideoLlama2 and Llava-NexT-Video achieved identical metrics. Qwen2-VL-7B-hf slightly exceeded its counterparts with higher METEOR scores. The improvements shown in both Table~\ref{tab:table02} and Table~\ref{tab:performance_improvements} indicate that the fine-tuning process effectively enhanced the models' capabilities across various question types.

\begin{table}[ht]
\centering
\caption{Performance improvements across models}
\label{tab:performance_improvements}
\begin{adjustbox}{width=0.46\textwidth}
\begin{tabular}{lccccc}
\toprule 
\midrule
\textbf{Model} & \textbf{Metric} & \textbf{Before} & \textbf{After} & \textbf{Improvement} \\
\midrule 
\multirow{5}{*}{VideoLLaMA2} 
 & BLEU   & 4.34 & 5.22 & \cellcolor{lightGreen}0.88 \\
 & ROUGE  & 27.63 & 28.30 & \cellcolor{lightGreen}0.67 \\
 & METEOR & 36.95 & 37.45 & \cellcolor{lightGreen}0.50 \\
 & CIDEr  & 1.41 & 1.54 & \cellcolor{lightGreen}0.13 \\
 & SPICE  & 25.03 & 26.51 & \cellcolor{mediumGreen}1.48 \\
\midrule
\multirow{5}{*}{Qwen2-VL-7B-Instruct} 
 & BLEU   & 0.99 & 7.40 & \cellcolor{darkGreen}6.41 \\
 & ROUGE  & 20.74 & 29.05 & \cellcolor{darkGreen}8.31 \\
 & METEOR & 36.60 & 38.17 & \cellcolor{mediumGreen}1.57 \\
 & CIDEr  & 1.12 & 1.71 & \cellcolor{lightGreen}0.59 \\
 & SPICE  & 14.60 & 23.54 & \cellcolor{darkGreen}8.94 \\
\midrule 
\multirow{5}{*}{LLaVA-NeXT-Video} 
 & BLEU   & 5.47 & 5.70 & \cellcolor{lightGreen}0.23\\
 & ROUGE  & 28.52 & 28.42 & \cellcolor{lightRed}-0.10 \\
 & METEOR & 37.70 & 38.25 & \cellcolor{lightGreen}0.55 \\
 & CIDEr  & 1.64 & 1.82 & \cellcolor{lightGreen}0.18 \\
 & SPICE  & 26.24 & 26.77 & \cellcolor{lightGreen}0.53 \\
\midrule 
\bottomrule
\end{tabular}
\end{adjustbox}
\end{table}

\subsection{Conceptual Evaluation Approach}

To conceptually evaluate the capabilities of VideoQA models in traffic intersection analysis, a series of experiments were conducted utilizing the InterAct VideoQA dataset. The primary focus was on assessing spatio-temporal reasoning, multi-agent interaction analysis, and counterfactual inference. The mathematical formalism for these assessments is described below.

\subsubsection{Spatio-Temporal Reasoning}
A model's ability to understand moving objects and their interactions over time can be captured using a temporal dependency function:

\begin{equation}
s_t = f(s_{t-1}, a_{t-1}, e_t)
\label{eq01}
\end{equation}

where:

$s_t$ is the state of the traffic scene at time $t$,

$s_{t-1}$ is the state at the previous time step,

$a_{t-1}$ represents the actions taken by vehicles/pedestrians at $t-1$,

$e_t$ denotes external factors such as traffic signals or environmental conditions.

The Equation \ref{eq01} models how the system state evolves dynamically over time, crucial for predicting future traffic behavior.

\subsubsection{Multi-Agent Interaction Analysis}
To quantify how different agents interact within a complex urban intersection a multi-agent dependency function is defined as:

\begin{equation}
I(A_i, A_j) = \sum_{t=1}^{T} \left( \frac{d(A_i, A_j)}{v_i + v_j + \epsilon} \right)
\label{eq02}
\end{equation}

where:

$I(A_i, A_j)$ represents the interaction score between two agents $A_i$ and $A_j$,

$d(A_i, A_j)$ is the Euclidean distance between the two agents at time $t$,

$v_i$ and $v_j$ are their respective velocities and

$\epsilon$ is a small constant to prevent division by zero.

A lower interaction score suggests a higher probability of collision or close interaction, which is crucial for identifying potential traffic incidents.

\subsubsection{Counterfactual Inference for Traffic Scenarios}
Counterfactual reasoning in AI-driven traffic analysis can be expressed using conditional probability as:

\begin{equation}
P(O | do(A = a')) \neq P(O | A = a)
\label{eq03}
\end{equation}

where:

$P(O | do(A = a'))$ represents the probability of an outcome $O$ occurring if action $A$ were forced to take value $a'$,

$P(O | A = a)$ is the observed probability of $O$ given the actual action $A = a$.

Equation \ref{eq03} captures the core of counterfactual traffic scenario analysis, allowing models to infer alternative outcomes had different traffic decisions have been made (e.g., \textit{“whether an incident could have been avoided if a driver had stopped at a red light”}).

\section{Discussion}
\label{discussion}

The InterAct VideoQA dataset presented in this paper represents a significant advancement in traffic intersection monitoring for VideoQA. Unlike many existing datasets that primarily focus on isolated events, object detection, or single-agent interactions, InterAct VideoQA captures the intricate and overlapping activities of real-world urban traffic. This complexity necessitates sophisticated reasoning capabilities, contextual awareness, and the ability to infer relationships between multiple entities over time. Evaluations using SOTA models such as VideoLLaMA2, Qwen2-VL-7B-Instruct, and LLaVA-NeXT-Video reveal both strengths and limitations in current VideoQA frameworks. While these models demonstrate competence in fundamental object recognition and basic event detection, they struggle with higher-order reasoning tasks that require understanding spatial relationships, sequential dependencies, and interactions among multiple agents. These shortcomings become particularly apparent in complex scenarios, such as determining the right-of-way in unmarked crossings, interpreting occluded objects in traffic congestion, or recognizing implicit causal relationships between consecutive events. However, fine-tuning these models with InterAct VideoQA enhanced their ability to process such intricate interactions, highlighting the dataset’s potential as a crucial resource for advancing VideoQA research in traffic monitoring.

One of the persistent challenges in VideoQA for traffic monitoring is robust spatio-temporal reasoning. Traffic events unfold dynamically over time, and answering questions such as \textit{“Did the blue sedan pass through the intersection before the pedestrian crossed?”} requires a precise understanding of temporal sequences and spatial positioning. Current models often struggle to maintain this level of detailed awareness across extended video frames, leading to errors in event ordering and causality inference. Addressing this issue requires improvements in long-term visual memory and attention mechanisms within VideoQA architectures. Another key challenge is multi-agent interaction analysis. Urban intersections feature a high degree of unpredictability, with interactions occurring between pedestrians, cyclists, and vehicles. Existing models frequently misinterpret interactions in dense and overlapping environments, leading to incorrect assessments of safety-critical situations, such as potential collisions or violations of traffic norms. Improving the ability of AI models to disentangle concurrent actions and differentiate between intentional movements and incidental proximity remains an open research problem.

Additionally, counterfactual reasoning presents a considerable obstacle. Many traffic-related VideoQA tasks involve hypothetical or alternative scenario-based questions, such as \textit{“Would the traffic have stopped if the signal had remained green?”} or \textit{“What would have happened if the cyclist had taken a different route?”} Current models, which predominantly rely on explicit visual cues and direct temporal progression, struggle with abstract reasoning and predictive modeling. Enhancing AI’s capacity for counterfactual inference requires integrating external knowledge sources, structured reasoning mechanisms, and causal learning frameworks. Overall, these evaluations underscore the strong need for advancements in spatio-temporal understanding, multi-agent analysis, and counterfactual inference, reinforcing the importance of richly annotated datasets to drive AI innovation forward.

\section{Conclusion}
\label{conclusion}
The study underscores the necessity for specialized VideoQA datasets to address the challenges of multi-event recognition in real-world traffic scenarios. InterAct VideoQA, designed for intersection-based traffic monitoring, provides annotated VideoQA pairs of concurrent interaction events. Experiments indicate that state-of-the-art VideoQA models require significant fine-tuning to accurately interpret complex traffic events, showing performance improvements with such adaptations. The findings highlight the need for end-to-end learning architectures tailored for dense, multi-event environments from live traffic feeds. To advance VideoQA models, InterAct VideoQA contributes to intelligent transportation research, enhancing traffic monitoring, urban mobility planning, and autonomous vehicle decision-making for safer and more efficient transportation systems. This dataset is envisioned as a long-term resource benefiting the research community, with plans to invite traffic data curators to contribute and expand it as an open-source dataset. To support wider accessibility and collaboration, InterAct VideoQA is now openly available to the public through a GitHub repository \url{https://github.com/joe-rabbit/InterAct_VideoQA}. \\

%%%%%%%%% REFERENCES

{\small
\bibliographystyle{ieee_fullname}
\bibliography{main}
}

\clearpage

\label{appendix} % --- LaTeX now numbers sections A, B, …

\section{Supplementary Material}
\subsection{Prompt For Generating QA's}

\begin{figure*}[t]
\centering
 \includegraphics[width=0.998\linewidth]{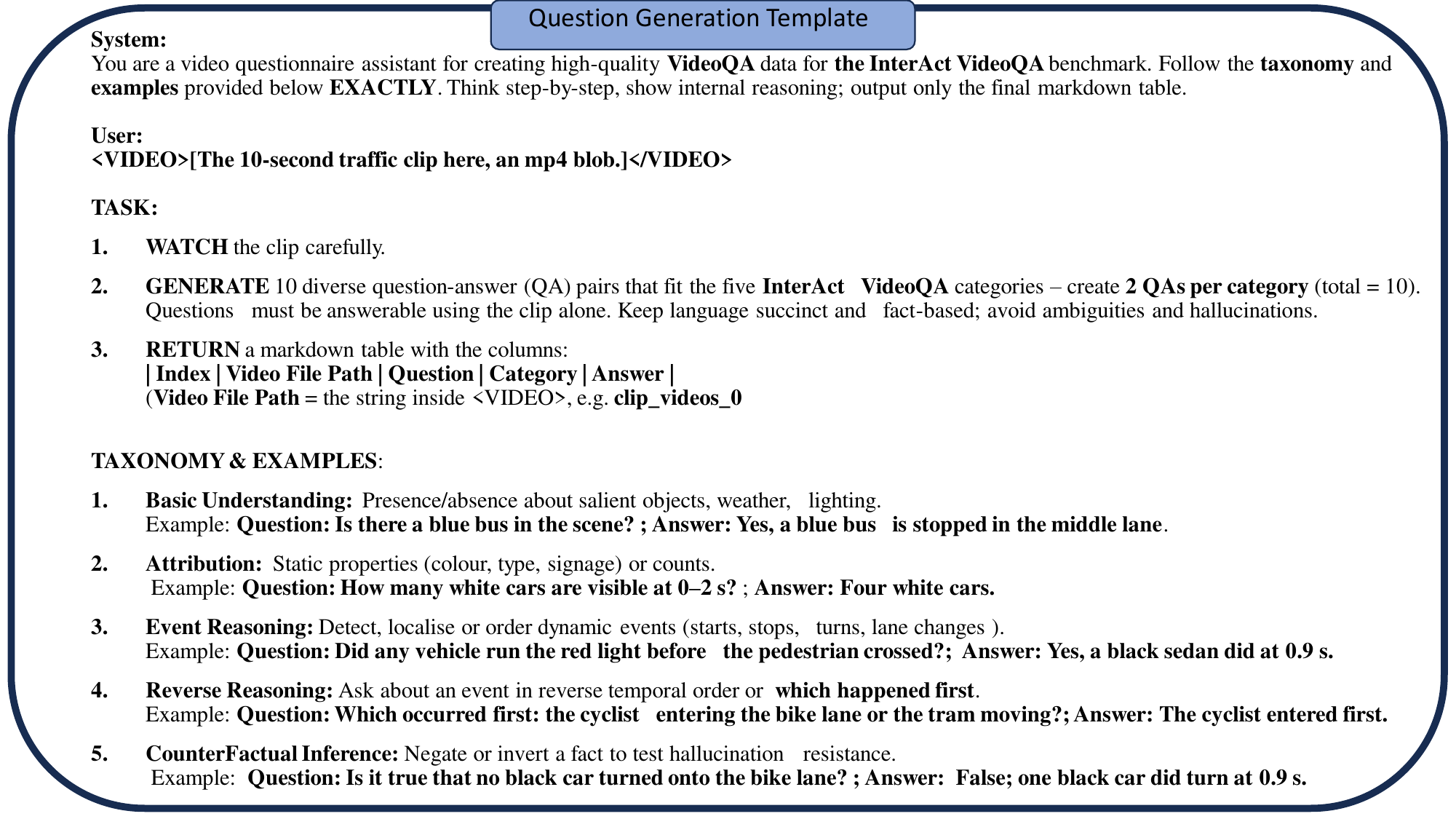}
\caption{Prompt used to generate InterAct VideoQA data.}
\label{fig:prompt}
\end{figure*}

This prompt fully specifies a controlled generation task for our VideoQA assistant.  First, it assigns the assistant the “video questionnaire” role and bounds it to the InterAct benchmark, ensuring consistency with our evaluation framework.  It then embeds a 10-second traffic clip (via an mp4 URL or blob) as the sole information source, preventing any external context or hallucination.  The task is broken into three clear steps: (1) watch the clip carefully, (2) produce exactly ten question–answer pairs, and (3) format the output as a Markdown table.  By mandating exactly two QAs per each of the five InterAct categories-basic Understanding, attribution, event reasoning, reverse reasoning, and counterfactual inference guarantees both coverage and diversity.  The prompt also enforces fact-based language and a one-sentence “Rationale” for each answer, which serves both as a correctness check and as training data for interpretability.  Finally, it prohibits revealing internal reasoning, shaping the model to output only the final structured table.

\subsection{Fine Tuning  parameters}
% ─── compact hyper-parameter table ──────────────────────────────────────────
Table \ref{tab:hyperparams_simple} summarizes the fine-tuning hyper-parameters used for all three backbone models. Each row lists the key optimization choices, epoch count, effective batch sizes, gradient accumulation steps, learning rate schedule, warm-up, weight decay, and optimizer, along with architectural notes (visual encoder freeze level and updated LLM layers). All models were adapted with LoRA rank-64 adapters, allowing full 7B–13B language backbones to be trained on $8$ × A100 GPUs within 6–11 GPU-hours while leaving the heavy visual encoders largely frozen. 
\begin{table*}[ht]
\centering
\caption{Fine-tuning hyperparameters for each model}
\label{tab:hyperparams_simple}
 \begin{adjustbox}{width=0.8\textwidth}
\begin{tabular}{l|ccc}
\toprule\toprule
\textbf{Hyper-parameter} &
\textbf{VideoLLaMA-2-7B-16F} &
\textbf{LLaVA-NeXT-Video-13B} &
\textbf{Qwen2-VL-7B-Instruct} \\
\midrule
Lora &
rank 64 &
rank 64 &
rank 64 \\[2pt] \cline{1-4} 
Global batch size & 2048 & 256 & 512 \\ \cline{1-4} 
Per-GPU batch      & 128  & 4   & 8   \\ \cline{1-4} 
Gradient accum.    & 4    & 8   & 8   \\[2pt] \cline{1-4} 
Peak LR            & \(2\times10^{-5}\) & \(2\times10^{-5}\) & \(1\times10^{-4}/5\times10^{-5}\) \\ \cline{1-4} 
LR schedule        & cosine & cosine & cosine \\ \cline{1-4} 
Warm-up            & 0.03 & 0.05 & 500 steps \\ \cline{1-4} 
Weight decay       & 0.01 & 0.01 & 0.01 \\[2pt] \cline{1-4} 
Visual encoder     & frozen ViT-CLIP & partial CLIP-14 & frozen Swin-V2 \\ \cline{1-4} 
LLM layers updated & all 7B & top 12 / 13B & full 7B \\ \cline{1-4} 
Optimizer          & AdamW & AdamW & AdamW \\  
\bottomrule\bottomrule
\end{tabular}
\end{adjustbox}
\end{table*}

Consistent AdamW optimization, cosine (or linear-to-cosine) LR decay, and a modest $0.01$ weight-decay coefficient were chosen to stabilize training across very different parameter scales, yielding reproducible improvements on every InterAct VideoQA split. All fine-tuning was carried out on the university's supercomputing resources.

\subsection{Ethics, Privacy, and Bias Mitigation}
\label{sec:ethics}

\subsubsection{Two-Stage Anonymisation Workflow}
\begin{enumerate}
  \item \textbf{Stage 1 – Automatic masking.}
        Each frame is tiled, super-resolved (\(\times2\) Real-ESRGAN) and fed to
        \emph{YOLOv8-small} (person class) and a fine-tuned
        licence-plate detector.
        \begin{itemize}
            \item Bodies are pixelated (\(6\times6\) blocks).
            \item Faces and plates are Gaussian-blurred (\(\sigma=21\) and
                  \(\sigma=31\), respectively).
            \item If no person is detected, \emph{MTCNN} is invoked to catch
                  tiny faces (\texttt{min\_face\_size}=12\,px).
        \end{itemize}
  \item \textbf{Stage 2 – Human audit.}
        Two annotators review the anonymised clip at \(1\times\) and
        \(0.25\times\) speed.
        Residual PII
        \emph{(i)} triggers re-processing or
        \emph{(ii)} causes permanent exclusion,
        subject to inter-rater agreement
        \(\kappa \ge 0.9\).
\end{enumerate}

\begin{table}[ht]
  \centering
  \caption{Key hyper-parameters for Stage 1 automatic masking.}
  \label{tab:privacy_hparams}
  \begin{adjustbox}{width=0.46\textwidth}
  \begin{tabular}{lcc}
    \toprule
    \textbf{Component} & \textbf{Setting} & \textbf{Reason} \\
    \midrule
    Tiling & \(640\times360\) (64\,px overlap) & retains \(<32\) px targets \\[2pt]
    Super-resolution & Real-ESRGAN, \(\times2\) & boosts micro-object recall \\[2pt]
    YOLOv8 conf.\ thresh. & 0.15 & lower bound for tiny objects \\[2pt]
    NMS IoU & 0.80 & keep neighbouring small boxes \\[2pt]
    Face blur & Gaussian, \(\sigma=21\) & GDPR “irreversible” blur \\[2pt]
    Body mask & Pixelate \(6\times6\) & harder to in-paint \\[2pt]
    Plate blur & Gaussian, \(\sigma=31\) & text removal \\[2pt]
    \bottomrule
  \end{tabular}
  \end{adjustbox}
\end{table}

% \subsubsection{Algorithmic description}

% \begin{algorithm}[ht]
% \caption{Frame-level anonymisation}\label{alg:mask}
% \begin{algorithmic}[1]
% \Require frame \(I\)
% \State \(T \gets \textsc{TileAndUpscale}(I)\) \Comment{tile + SR}
% \State \(B_{\mathrm{person}},B_{\mathrm{plate}} \gets \varnothing\)
% \ForAll{tile \(t\in T\)}
%     \State \(B_{\mathrm{person}} \gets B_{\mathrm{person}}\cup
%            \textsc{YOLOv8}(t,\textsc{person})\)
%     \State \(B_{\mathrm{plate}} \gets B_{\mathrm{plate}}\cup
%            \textsc{YOLOv8\_LP}(t)\)
% \EndFor
% \If{\(B_{\mathrm{person}} = \varnothing\)}
%     \State \(B_{\mathrm{face}} \gets \textsc{MTCNN}(I)\)
% \Else
%     \State \(B_{\mathrm{face}} \gets \varnothing\)
% \EndIf
% \ForAll{box \(b\in B_{\mathrm{person}}\)} \State \textsc{Pixelate}(I,b) \EndFor
% \ForAll{box \(b\in B_{\mathrm{plate}}\cup B_{\mathrm{face}}\)}
%     \State \textsc{GaussianBlur}(I,b)
% \EndFor
% \State \Return masked frame \(I\)
% \end{algorithmic}
% \end{algorithm}

\subsubsection{Bias-Aware QA Generation}

During GPT-driven question generation we compute an
\emph{FCA} (Fairness–Consistency–Accuracy) score:
\[
\text{FCA}=\tfrac13\bigl(F_{\text{div}}+C_{\text{dup}}+A_{\text{exact}}\bigr)
\]
Batches with \(\text{FCA}<0.70\) are automatically rejected for rewriting,
ensuring balanced class coverage, low duplication, and answer correctness.

\subsubsection{Controlled Dataset Release}

Researchers request access via a Google Form
(e-mail, ORCID, purpose, license agreement).
Requests are reviewed weekly, and approved users receive a presigned URL.

\clearpage
\subsection{InterAct VideoQA Dataset Samples}

\begin{figure*}[b]
  \centering
  \includegraphics[width=0.95\linewidth]{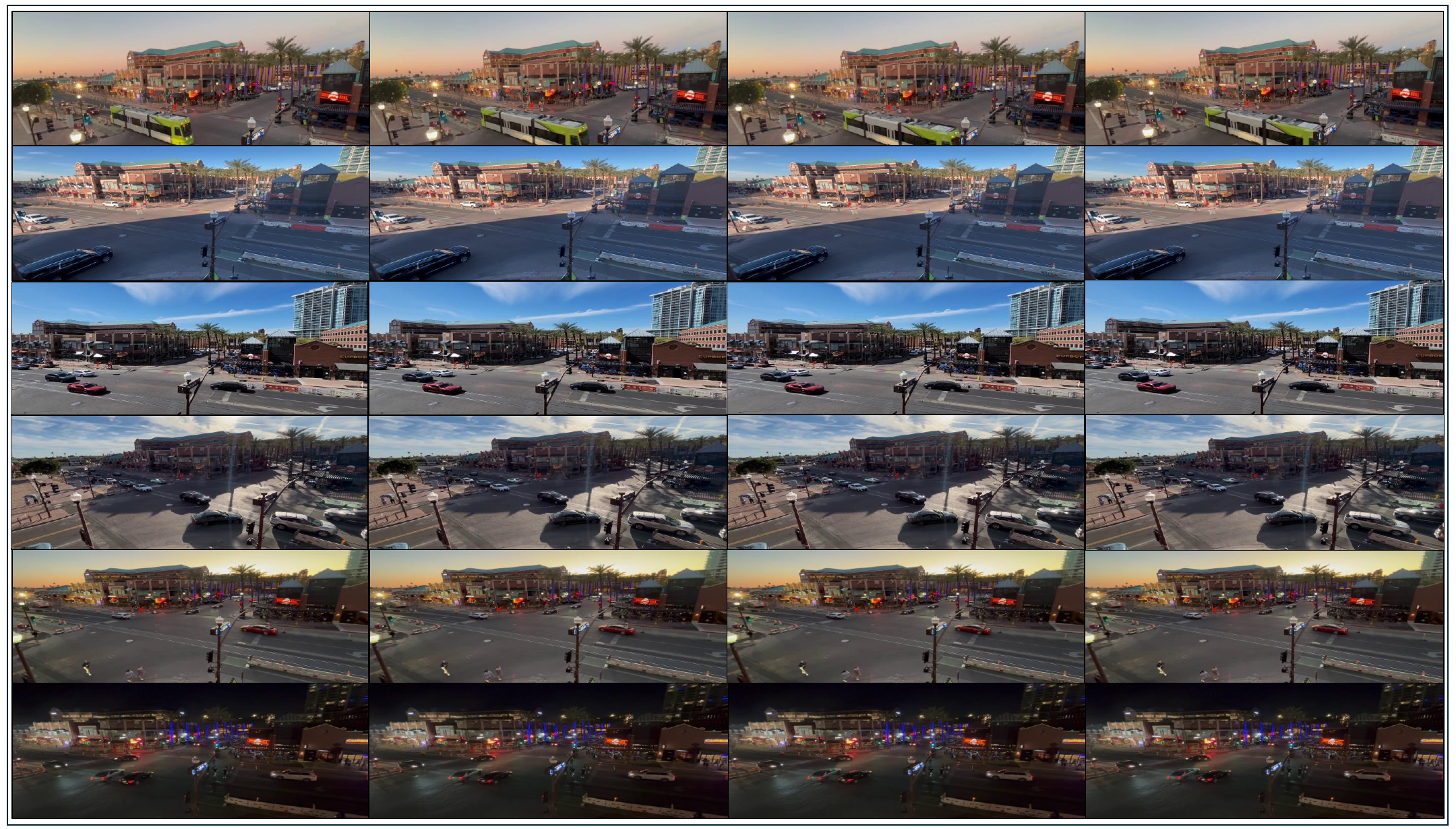}
  \caption{Lighting Conditions - Frames illustrate varying times of day: (top) sunrise (6:40 AM), (second) 7:30 AM, (third) 12:30 PM, (fourth) 4:00 PM, (fifth) sunset (6:30 PM), and (last) night (8:00 PM), ensuring diverse lighting coverage.}
  \label{A1}
\end{figure*}

\begin{figure*}[b]
  \centering
  \includegraphics[width=0.95\linewidth]{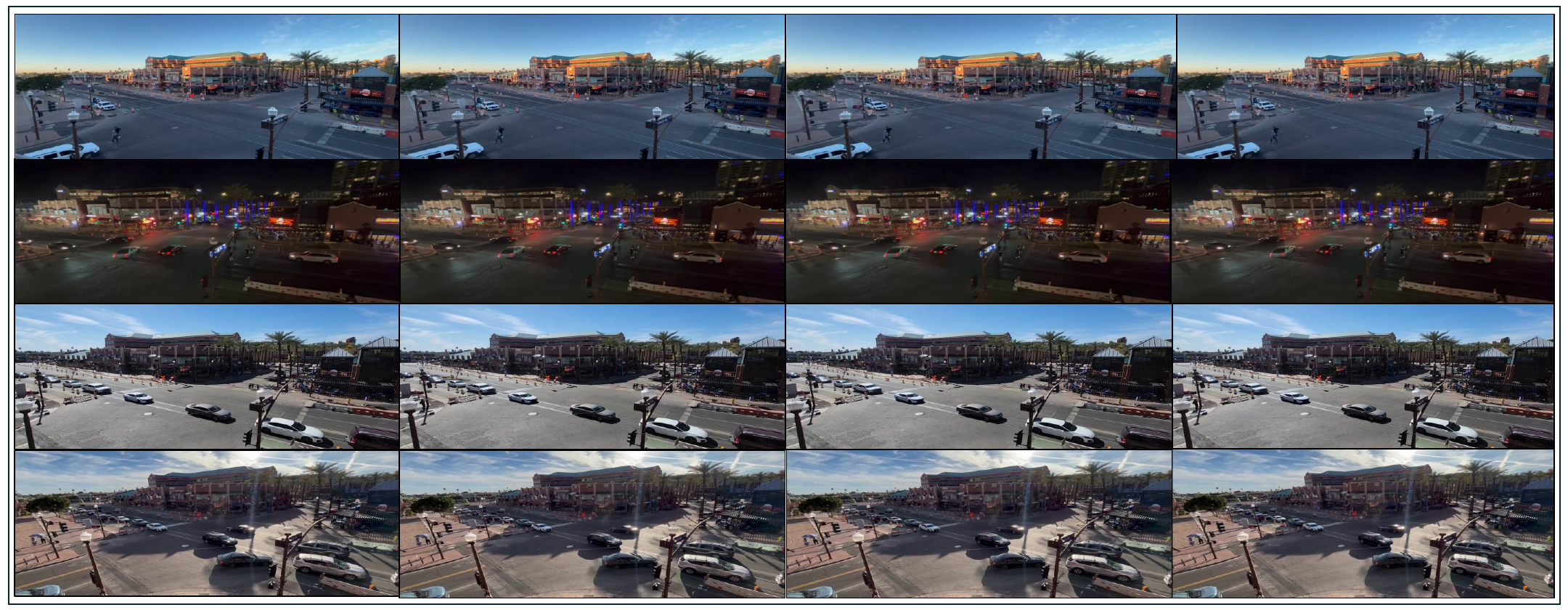}
  \caption{Traffic Density Analysis - The first row (7:00 AM) shows minimal movement, the second (6:30 PM) captures light traffic, the third (noon) shows moderate congestion in one direction, and the last (4:30 PM) depicts peak traffic with heavy congestion and increased interactions.}
  \label{A2}
\end{figure*}

\begin{figure*}[b]
  \centering
  \includegraphics[width=0.95\linewidth]{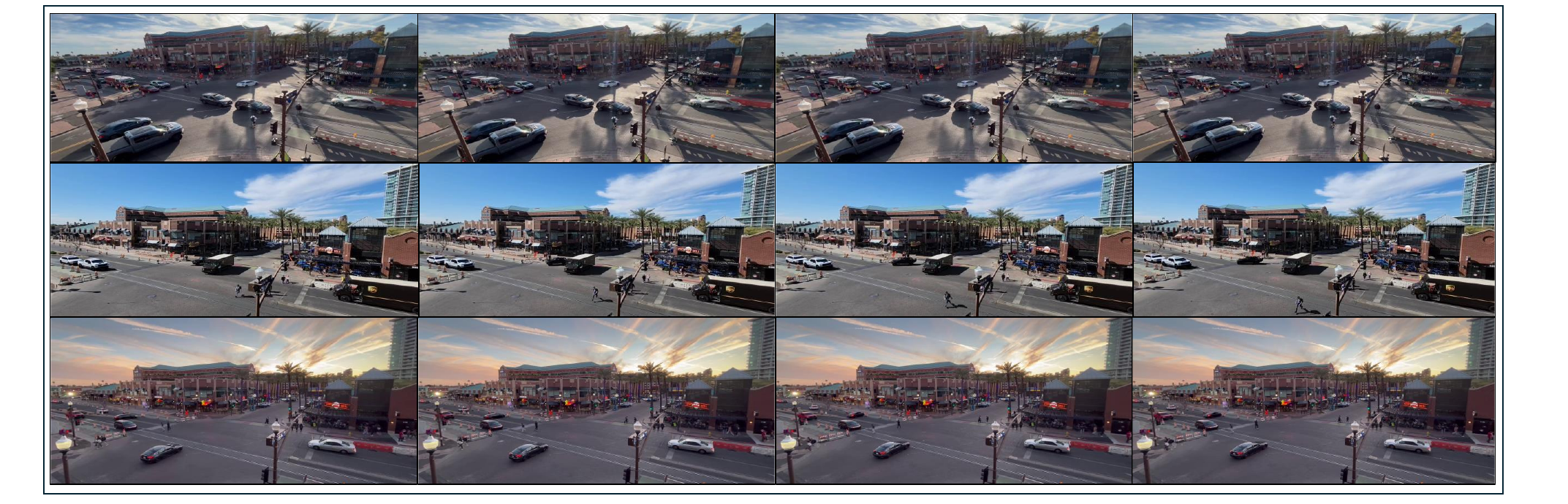}
  \caption{Complex Interactions – The first row (4:30 PM) captures a skateboarder crossing against a red light, momentarily yielding to a car on the green. The second row (noon) shows a pedestrian running to cross before vehicles start moving. The last row (6:30 PM) features a black sedan executing a U-turn and briefly blocking traffic. These scenarios underscore the dataset’s ability to capture unique, real-world vehicle-pedestrian interactions and complex traffic behaviors.}
  \label{A3}
\end{figure*}

\clearpage
\FloatBarrier % Ensures all previous figures are placed before the new subsection
\subsubsection{Question-Answers Samples of the Dataset}

\begin{figure*}[t]
  \centering 

\includegraphics[width=0.95\linewidth]{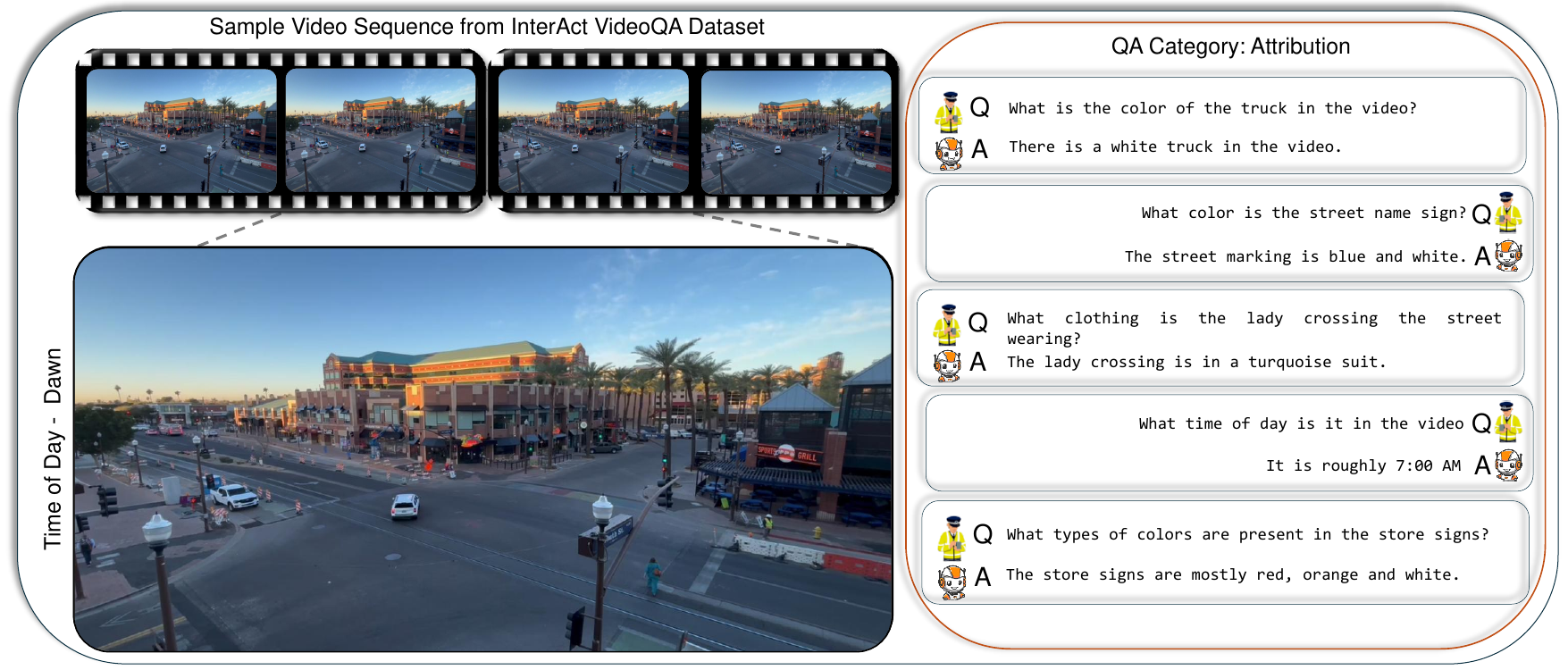}
  \caption{Dawn - The figure illustrates question samples for pictures captured during dawn (6:30 AM) at the intersection. And all questions of attribution are associated with questions such as what is the color of the truck in the video. Images such as the questions include describing the vehicles, street signs, clothes of pedestrians, and colors present on the store sign.}
  \label{A4}
\end{figure*}

\begin{figure*}[b]
  \centering
\includegraphics[width=0.95\linewidth]{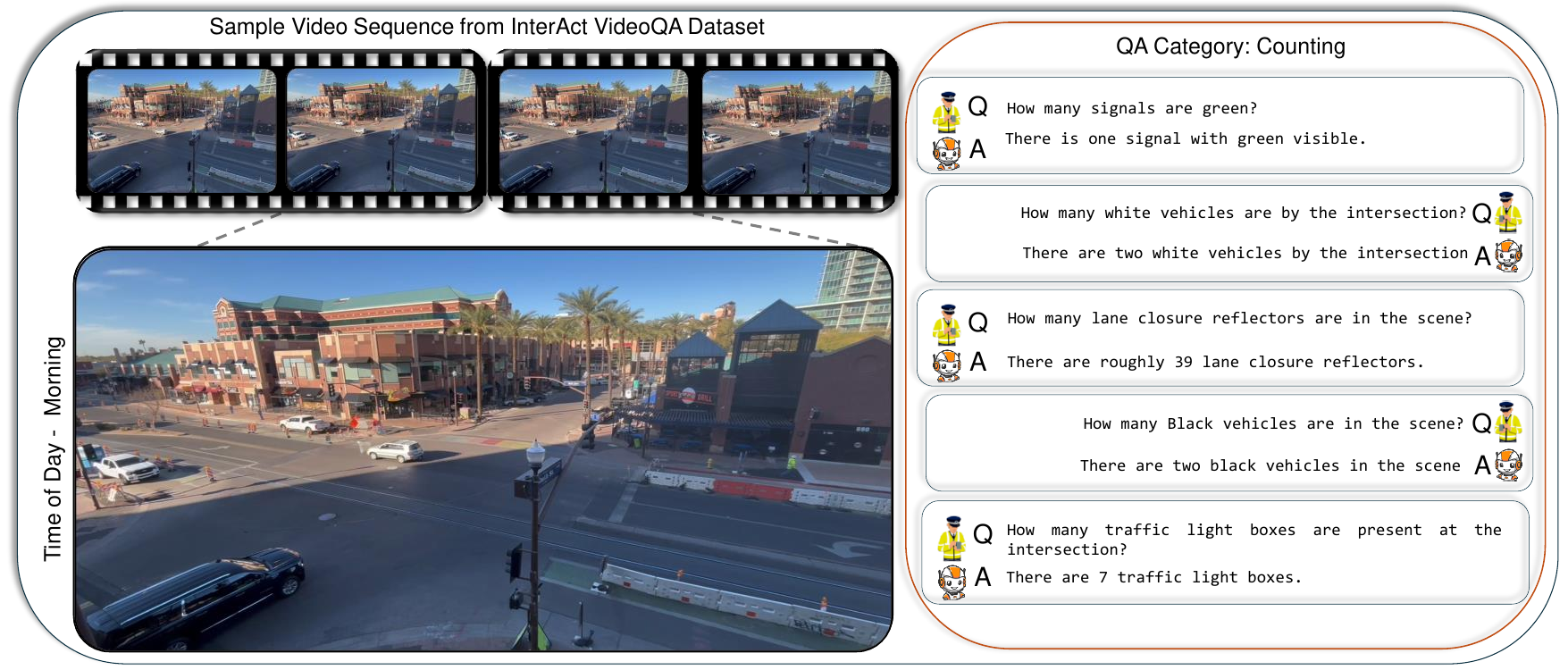}
\caption{Morning - The figure illustrates question samples for pictures captured at 8:00 AM at the intersection. The questions based on counting are associated with the video such as how many green signals are visible how many intersections how many reflectors and black vehicles are present how many traffic light boxes are present.}
\label{A5}
\end{figure*}

\begin{figure*}[b]
  \centering
\includegraphics[width=0.95\linewidth]{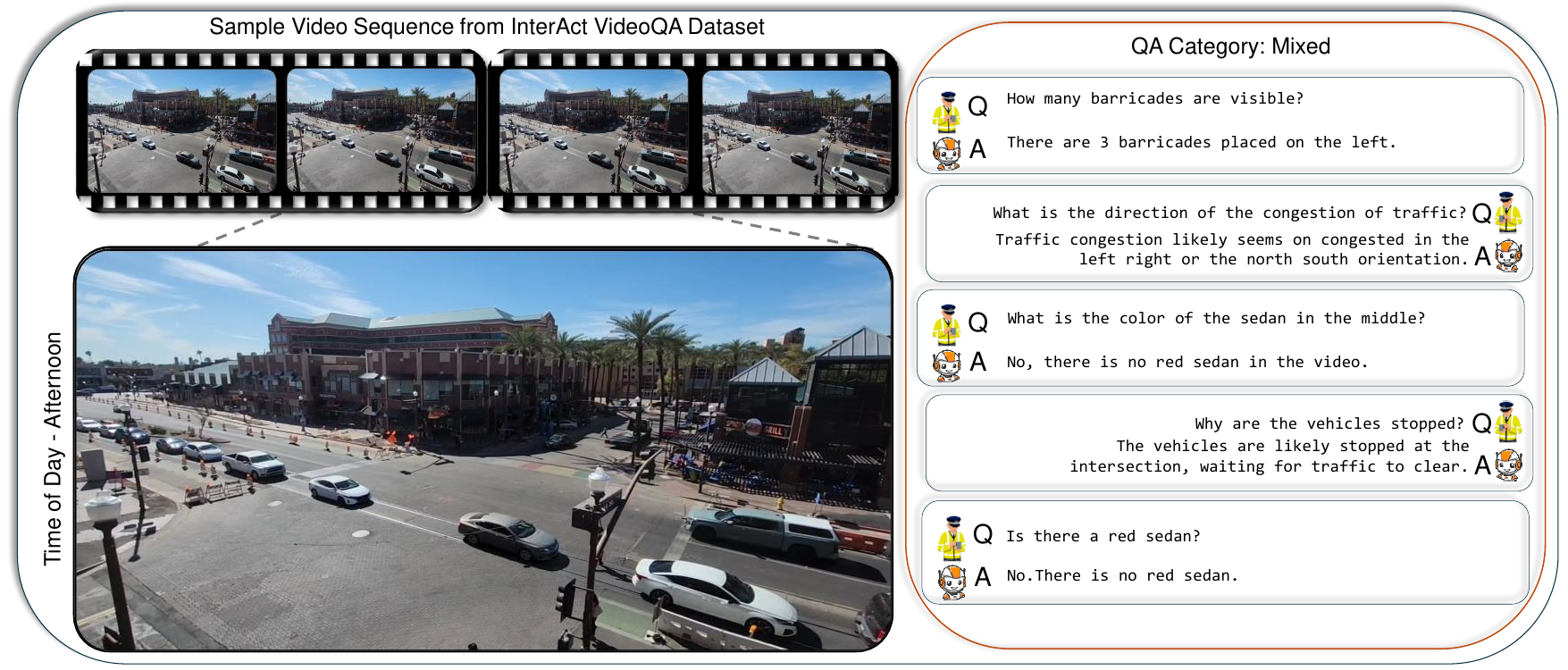}
\caption{Afternoon - The figure illustrates more complex, multi-faceted questions that combine object counting, traffic flow analysis, and counterfactual reasoning. The questions evaluate the understanding of traffic congestion patterns, infrastructure elements like barricades, and vehicle presence verification, demonstrating the model's ability to handle compound reasoning tasks.}
\label{Aafternoon}
\end{figure*}

\begin{figure*}[b]
  \centering
\includegraphics[width=0.95\linewidth]{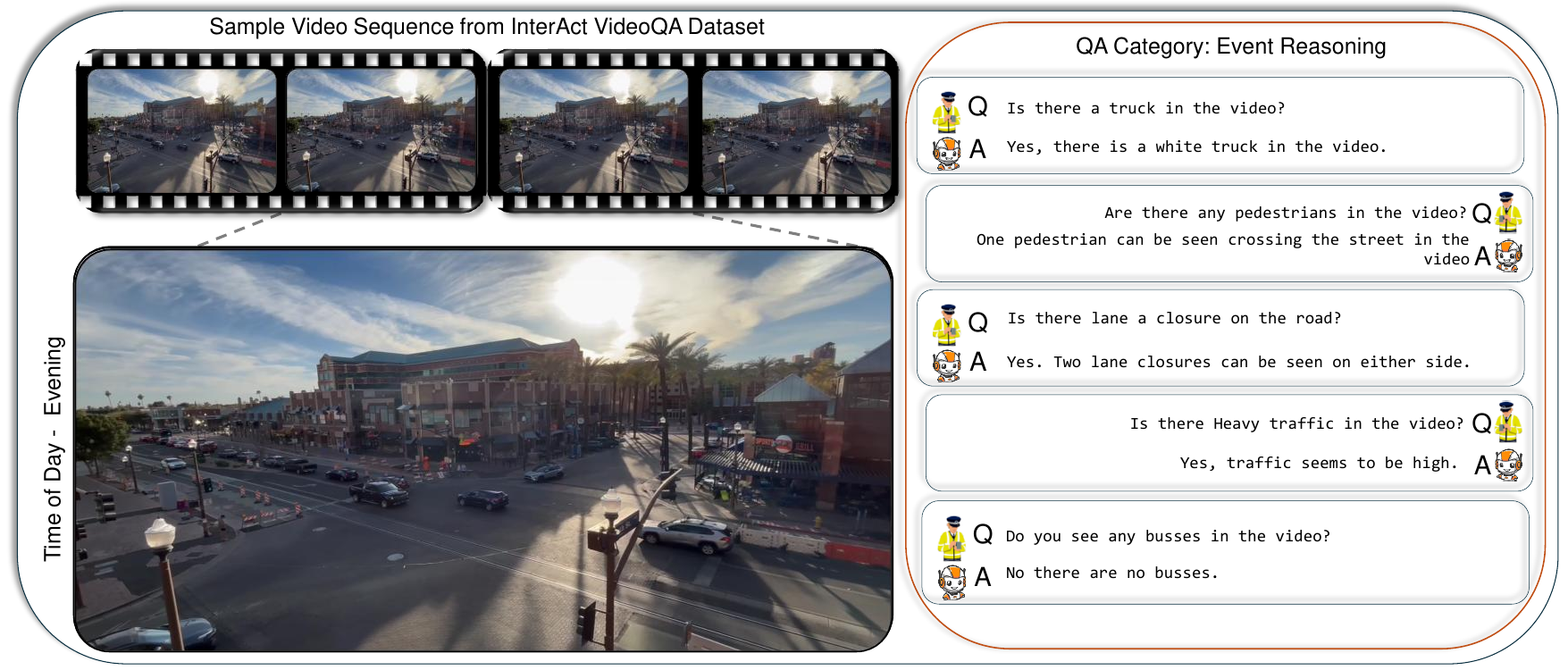}
\caption{Evening - The figure illustrates event-based reasoning questions. The questions focus on the presence of specific objects (e.g., trucks, buses), pedestrian movement, lane closures, and traffic density.}
\label{A6}
\end{figure*}

\begin{figure*}[b]
  \centering
\includegraphics[width=0.95\linewidth]{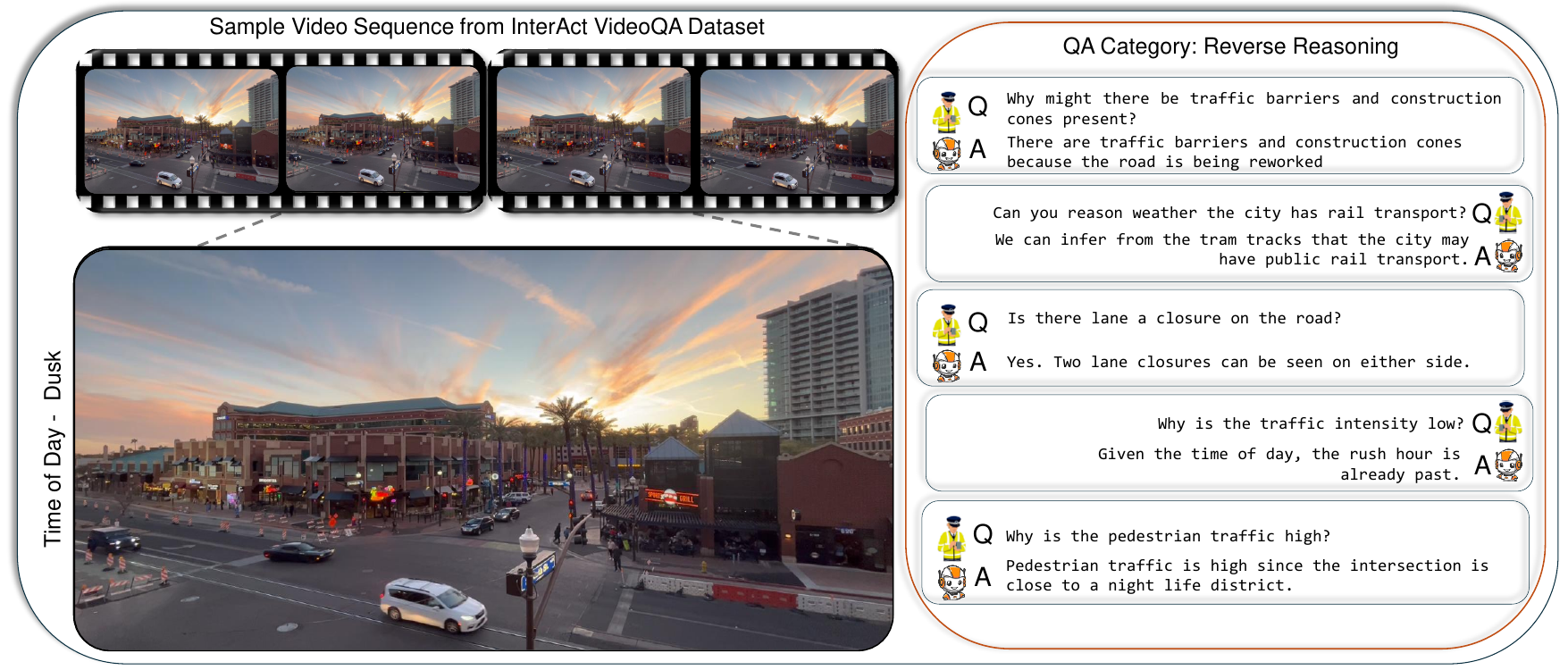}
\caption{Dusk - The figure illustrates a dusk scenario where reverse reasoning questions are explored. The QA set examines logical deductions, such as the presence of construction barriers, public transport inference, lane closures, and pedestrian traffic intensity.}
\label{A7}
\end{figure*}

\begin{figure*}[b]
  \centering
\includegraphics[width=0.95\linewidth]{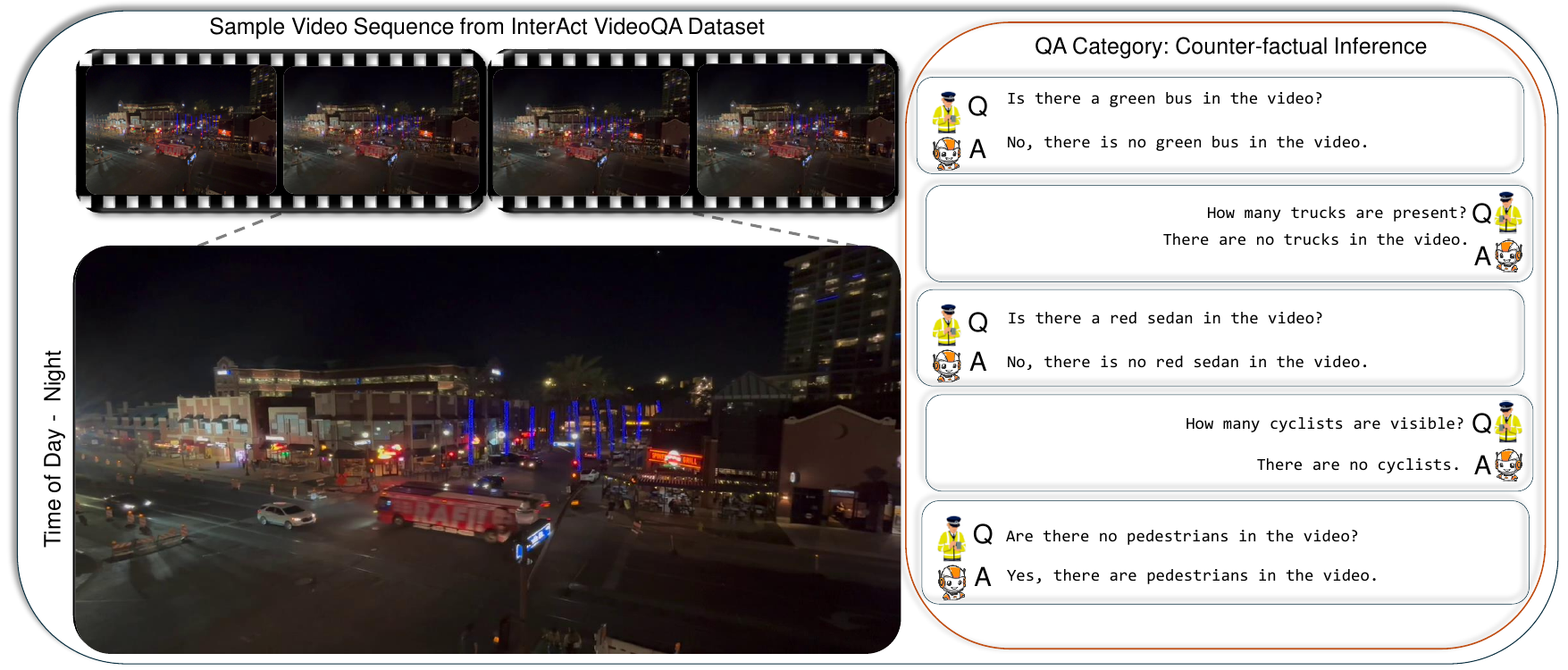}
\caption{Night - The figure illustrates counterfactual inferencing, designed to test the model's ability to avoid hallucinations in low-light conditions. These questions focus on verifying the presence or absence of various road users such as vehicles, cyclists, and pedestrians, ensuring reliable performance in challenging visibility conditions.
}
\label{A8}
\end{figure*}

\end{document}